\pdfoutput=1

\documentclass[11pt]{article}

\usepackage[]{acl}

\usepackage{times}
\usepackage{latexsym}

\usepackage[T1]{fontenc}

\usepackage[utf8]{inputenc}

\usepackage{microtype}

\usepackage{inconsolata}

\usepackage{graphicx}

\usepackage{bm}
\usepackage{amsmath}
\usepackage{url}
\usepackage{graphicx}
\usepackage{float} 
\usepackage{subfigure}
\usepackage{multirow}
\usepackage{makecell}
\usepackage{xpinyin}
\usepackage{enumerate}

\usepackage{xcolor}
\usepackage{bbding}
\usepackage{soul}
\newcolumntype{L}[1]{>{\raggedright\let\newline\\\arraybackslash\hspace{0pt}}m{#1}}
\setul{2pt}{2pt}
\usepackage{booktabs}
\usepackage{amsmath,amsfonts}
\usepackage[utf8]{inputenc}
\usepackage{cleveref}

\crefname{section}{§}{§§}
\Crefname{section}{§}{§§}
\usepackage{enumitem}
\setenumerate[1]{itemsep=0pt,partopsep=0pt,parsep=\parskip,topsep=5pt}
\setitemize[1]{itemsep=0pt,partopsep=0pt,parsep=\parskip,topsep=5pt}
\setdescription{itemsep=0pt,partopsep=0pt,parsep=\parskip,topsep=5pt}
\usepackage{arydshln}

\usepackage[linesnumbered,ruled,vlined]{algorithm2e}

\title{SlangDIT: Benchmarking LLMs in Interpretative Slang Translation}

\author{Yunlong Liang, \ Fandong Meng\thanks{ \ \ Corresponding author.},  \ Jiaan Wang, \ Jie Zhou \\
Pattern Recognition Center, WeChat AI, Tencent Inc \\ 
\texttt{\{yunlonliang,fandongmeng,torchwang,withtomzhou\}@tencent.com} \\
}

\begin{document}
\maketitle
\begin{abstract}
The challenge of slang translation lies in capturing context-dependent semantic extensions, as slang terms often convey meanings beyond their literal interpretation. While slang detection, explanation, and translation have been studied as isolated tasks in the era of large language models (LLMs), their intrinsic interdependence remains underexplored. The main reason is lacking of a benchmark where the two tasks can be a prerequisite for the third one, which can facilitate idiomatic translation. In this paper, we introduce the interpretative slang translation task (named SlangDIT) consisting of three sub-tasks: slang detection, cross-lingual slang explanation, and slang translation within the current context, aiming to generate more accurate translation with the help of slang detection and slang explanation. To this end, we construct a SlangDIT dataset, containing over 25k English-Chinese sentence pairs. Each source sentence mentions at least one slang term and is labeled with corresponding cross-lingual slang explanation. Based on the benchmark, we propose a deep thinking model, named SlangOWL. It firstly identifies whether the sentence contains a slang, and then judges whether the slang is polysemous and analyze its possible meaning. Further, the SlangOWL provides the best explanation of the slang term targeting on the current context. Finally, according to the whole thought, the SlangOWL offers a suitable translation. Our experiments on LLMs (\emph{e.g.}, Qwen2.5 and LLama-3.1), show that our deep thinking approach indeed enhances the performance of LLMs where the proposed SLangOWL significantly surpasses the vanilla models and supervised fine-tuned models without thinking.~\footnote{https://github.com/XL2248/SlangDIT} 

\end{abstract}

\section{Introduction}
The slang includes words, phrases, idioms, and expressions that are not typically found in formal language or standard dictionaries, conveying subtle shades of meaning, tone, and attitude, which is often used in social interactions, particularly among specific groups or communities~\cite{mashhady2013slang}. The natural characteristic leads to some challenges of slang translation since it is hard to model context-dependent semantic extensions. Although plenty of studies on them have been carried out based on slang detection~\cite{ishiwatari-etal-2019-learning,pei-etal-2019-slang}, or cross-lingual slang explanation~\cite{gluck2025clixcrosslingualexplanationsidiomatic}, or slang translation~\cite{sun-etal-2022-semantically} in the era of large language models (LLMs)~\cite{jhirad-etal-2023-evaluating,pei-etal-2019-slang,sun-etal-2022-semantically,sun-etal-2024-toward}, to our knowledge, little research work has been devoted to slang translation with the help of slang detection and cross-lingual slang explanation. One important reason is the lack of such slang translation datasets.

\textbf{\begin{figure}[t]
    \centering
    \includegraphics[width=0.5\textwidth]{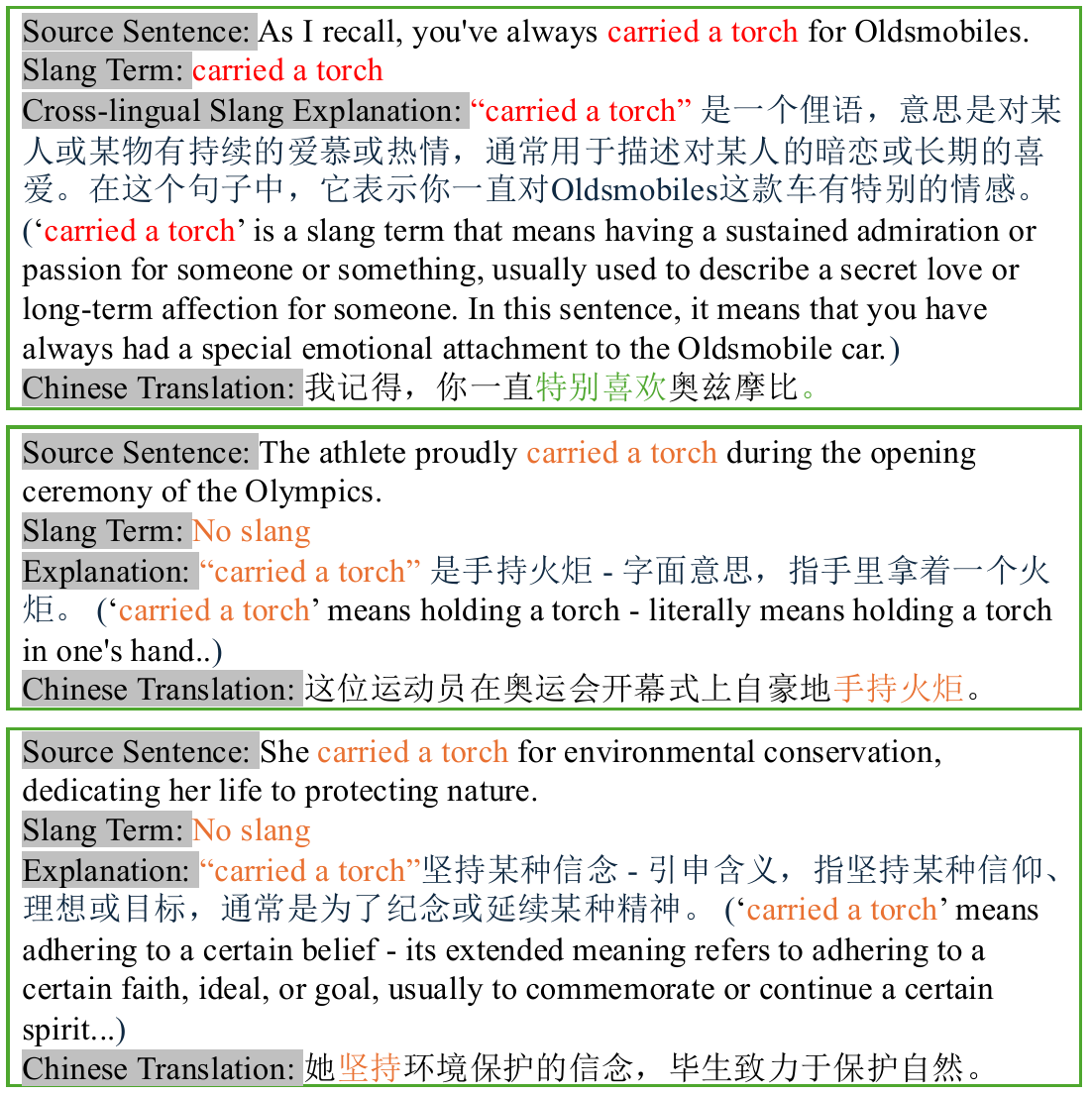}
    \caption{Some examples of the SlangDIT benchmark.
    }
    \label{fig.1}\vspace{-10pt}
\end{figure}}
Meanwhile, the previous work generally neglects the polysemy present in slang terms. In different contexts, the slang term conveys different ideas. For example, in the Figure~\ref{fig.1}, the term `\emph{carried a torch}' is a slang term in the top box while it is not in the middle (`superficial semantic meaning') and bottom box (`extended meaning'). In fact, for humans, to identify and understand slang, one firstly needs to be familiar with the cultural, social, and historical context in which it is used. This involves recognizing the nuances of language and also requires an awareness of the ever-changing nature of language, as slang terms and expressions may quickly evolve or become outdated~\cite{legaudaite2010understanding,keidar-etal-2022-slangvolution}. For instance, if we take the slang term (\emph{i.e.} `\emph{carried a torch}') as general words in the top box of Figure~\ref{fig.1}, we could not capture the real sense the speaker said (\emph{i.e.}, `sustained admiration or passion for
someone or something'). Secondly, to translate it, one requires not only conveying the literal meaning of the words but also capturing the tone, connotation, and implied meaning that is often embedded in slang expressions~\cite{mattiello2009difficulty}. Even for professional human translators, they sometimes fail to convey the intended meaning in practice. As shown in the example of Figure~\ref{fig.1}, directly translate the slang term can not show the subtle meaning the sentence reflects. And with the help of the slang detection and cross-lingual slang explanation, the translation will express the original intention and become satisfactory. All of the above call for a such data resource that can encourage further research in slang understanding and translation. 

In this work, we propose a new task named \textbf{I}nterpretative \textbf{Slang} \textbf{T}ranslation (SlangDIT), with the goal to produce more accurate translations by taking the detected slang and the corresponding cross-lingual slang explanation. To this end, we firstly construct a SlangDIT dataset. Specifically, based on the large-scale movie subtitles\footnote{\url{https://www.jubenz.com/}}, about 28M English-Chinese sentence pairs~\cite{liang-etal-2022-msctd}, (1) we use advanced LLMs (\emph{e.g.}, Qwen2.5-72b~\cite{qwen2.5}) to judge whether the English sentence contains a slang. (2) for the sentence with a slang (\textasciitilde 776k), we further utilize Qwen2.5-72b and Llama3.3-70b to extract the slang terms. To ensure the data quality, we only maintain the sentence where both of Qwen2.5-72b and Llama3.3-70b predict the same slang term and GPT-4o agrees that it is a slang. (3) We further utilize Qwen2.5-72b produce the Chinese explanation. (4) To explore the context impact on the slang term, we utilize GPT-4o to annotate whether each slang term is polysemous. Consequently, we obtain over 25k English-Chinese sentences with 13,580 genenal slang terms and 7,818 polysemous slang terms. 

Based on the constructed SlangDIT dataset, we propose a deep thinking model (SlangOWL) that simulates cognitive process of humans. Specifically, the model firstly identifies whether the sentence contains a slang, and then judges whether the slang is polysemous and analyze its possible meaning. Further, it provides the best cross-lingual explanation of the slang term targeting on the current context. Finally, according to the thought, the model offers a suitable translation. To achieve this goal, we need the long thought samples to train our models. In view of promising reasoning ability in existing o1-like LLMs, we decide to provide the four key elements (slang term, polysemy, cross-lingual explanation and translation) to DeepSeek-R1-Distill-Qwen-32B~\cite{deepseekai2025deepseekr1incentivizingreasoningcapability}, and collect the thought process. 

Experiments on three LLM-based systems~\cite{qwen2.5,dubey2024llama}, \emph{i.e.}, Qwen2.5-7B-Instruct, Llama-3.1-8B-Instruct and Qwen2.5-14B-Instruct, show the effectiveness of deep thinking on translation. It significantly outperforms the vanilla models and simply supervised fine-tuned models in terms of BLEU~\cite{papineni-etal-2002-bleu}, CometKivi~\cite{rei-etal-2022-cometkiwi}, Comet~\cite{rei-etal-2020-comet} and evaluators via GPT-4o. 

In summary, our main contributions are:
\begin{itemize}[leftmargin=*,topsep=0pt]

\item We propose a new task: interpretative slang translation named SlangDIT, consisting of three sub-tasks, to advance slang understanding and translation research.

\item We are the first that contributes the SlangDIT dataset, which contains 25k <English sentence, Chinese sentence, slang terms, cross-lingual slang explanation> quadruplets. Particularly, it offers 7,818 polysemous slang terms.

\item We propose a SlangOWL model that achieves the best performance with the help of deep thinking during translation. We also show that the slang detection and cross-lingual explanation play a key role in translating the sentence with a slang term.

\end{itemize}

\begin{table*}[htbp]
\centering
\setlength{\tabcolsep}{0.9mm}{
\begin{tabular}{clrrrrccrrr}
\toprule
&\multirow{2}{*}{\bf{Type}} &\multicolumn{3}{c}{\#\bf{Sentences}} &\multicolumn{2}{c}{\# \bf{Slang Type}} &\multirow{2}{*}{\bf{XLSE}} &\multirow{2}{*}{\#\bf{AvgEn}} &\multirow{2}{*}{\#\bf{AvgZh}} &\multirow{2}{*}{ \#\bf{AvgExp} }\\

\cmidrule(r){3-5} \cmidrule(r){6-7} 
&&No & \emph{w/} NPST &\emph{w/} PST & NPST &PST & & & \\

\midrule

\multirow{4}{*}{{\bf{SlangDIT}}} 
&Train    &39,980 &13,580 &26,400 &13,580 &6,653 &39,980&7.89 &12.47 &74.06 \\
&Valid    &1,815 &1,815 &0 &1,815 &0 &1,815&7.93 &12.71 &76.65 \\
&Test     &1,863 &1,863 &0 &1,863 &0 &1,863&7.88&12.77 &77.12 \\
&Hard Test     &0 &0 &1,165 &0 &1,165 &1,165 &7.86&12.57 &75.54 \\

\bottomrule
\end{tabular}}
\caption{Detailed Statistics of our SlangDIT dataset. `NPST' and `PST' means non-polysemous slang terms and polysemous slang terms; `XLSE' means cross-lingual slang explanation; \#: number of the corresponding item, \emph{i.e.}, AvgEn: Average length of each utterance in English (word level); AvgZh/AvgExp: Average length of each sentence in Chinese (character level).}
\label{Tbl:data}\vspace{-10pt}
\end{table*}

\section{SlangDIT Task}
\label{sec:td}
In this section, we firstly clarify the symbol definition, and then define the proposed~\emph{Interpretative Slang Translation} task. 

Given an input sentence in the source language $X = X_1, X_2, X_3,..., S_1, S_2,..., S_p,..., X_m$ where the $X_i$ is the token and the $S_i$ is the token belongs to a slang, the goal of the SlangDIT task is to identify the slang term $S = Y^s_{1},Y^s_{2},Y^s_{3},..., Y^s_p$, and then generate its explanation in a target language $E = Y^e_1, Y^e_2, Y^e_3, ,...,Y^e_k$, and finally output its translation in a target language $Y = Y^t_1, Y^t_2, Y^t_3,...,Y^t_m$.

Formally, the probability distribution of the target output $S, E, Y$ are defined as follows:
\begin{equation}
\label{eq:dnmt}
        {P}(S, E, Y|X) = \prod_{r=1}^{R}p(Y^{*}_{r}|Y^{*}_{<r}, X),
\end{equation}
where ${*} \in \{s, e, t\}$ and $Y_{<r} = \{Y_{1}, Y_{2}, Y_{3}, ..., Y_{r-1}\}$ and $R = p + k + m$.

\section{SlangDIT Dataset}
In this section, we mainly introduce our SlangDIT dataset in five aspects: \emph{Data Source}~\autoref{ds}, \emph{Annotation Procedure}~\autoref{ap}, \emph{Annotation Quality Assessment}~\autoref{aqa}, \emph{Dataset Statistics}~\autoref{dsa}, and the introduction of \emph{Related Datasets}~\autoref{rds}.

\subsection{Data Source}
\label{ds}
Because movie subtitles contain utterances that better reflect natural conversations which usually involve slang terms~\cite{sun-etal-2024-toward,chen2024largelanguagemodelsclassical}, we thus choose the movie subtitle as our data source, \emph{e.g.}, the large-scale movie subtitles~\cite{liang-etal-2022-msctd} and OpenSubtitles~\cite{lison-tiedemann-2016-opensubtitles2016}\footnote{\url{http://www.opensubtitles.org/}}. Due to the dataset of~\citet{liang-etal-2022-msctd} offers the corresponding Chinese translation, we take this dataset as our choice. However, the lack of associated slang annotation and cross-lingual slang explanation makes it impossible for directly conducting research on interpretative slang translation. Therefore, we further annotate slang term and the corresponding slang term explanation.

\begin{table*}[ht!]
\centering
\scalebox{0.72}{
\setlength{\tabcolsep}{0.90mm}{
\begin{tabular}{lccccccccc}
\toprule
\multirow{2}{*}{\textbf{Datasets}} &\multirow{2}{*}{\textbf{SD}} &\multirow{2}{*}{\textbf{XLSE}}&\multirow{2}{*}{\textbf{Translation}} &\multirow{2}{*}{\textbf{Polysemy}}&  \multicolumn{3}{c}{\textbf{\#Slang Terms}}  \\
\cmidrule(r){6-8} 
 & & & &&Train & Valid & Test\\
\midrule
{Urban Dictionary}~\cite{ni-wang-2017-learning}   & \Checkmark  & \XSolidBrush   &\XSolidBrush & \XSolidBrush & 371,028  &- &50,000   \\
{OSD}~\cite{sun-etal-2022-semantically}   & \Checkmark  & \XSolidBrush   &\XSolidBrush  & \XSolidBrush &1,635  & - &299   \\

{GDoS}~\cite{adams2012green}   & \Checkmark  & \XSolidBrush   &\XSolidBrush  & \XSolidBrush &-  &- &-   \\

{CLIX}~\cite{gluck2025clixcrosslingualexplanationsidiomatic}   &\XSolidBrush  & \Checkmark (English$\rightarrow$Spanish/German)  &\XSolidBrush & \XSolidBrush &278 &150 &200   \\

{OpenSubtitles-Slang}~\cite{sun-etal-2024-toward}   & \Checkmark  & \XSolidBrush (English$\rightarrow$English)  &\Checkmark(English$\rightarrow$German/French) & \XSolidBrush &836  &- &-   \\\cdashline{1-8}[4pt/2pt]

SlangDIT (Ours)   &\Checkmark & \Checkmark (English$\rightarrow$Chinese)  &\Checkmark (English$\rightarrow$Chinese)  &\Checkmark  &20,233  &1,815 &3,028   \\
\bottomrule
\end{tabular}}}
\caption{Comparison of previous slang detection dataset: Urban Dictionary, OSD, GDoS, and OpenSubtitles-Slang; (2) cross-lingual slang explanation datasets: CLIX and OpenSubtitles-Slang, and (3) our SlangDIT. `SD' means slang detection and `polysemy' means polysemous labeling. } 
\label{data:tr}
\end{table*}

\subsection{Annotation Procedure}
\label{ap}

Since the full data are large (\textasciitilde28M), the annotation procedure is automatic to build the SlangDIT dataset via advanced LLMs (Qwen2.5-72b, Llama3.3-70b and GPT-4o), which includes four steps: slang judging, slang extraction, explanation generation and polysemy annotation~\footnote{The prompt used in this process are presented in Figure~\ref{annotaion_app} \textasciitilde ~\ref{annotaion_app4} of Appendix.}. 

\noindent\textbf{Slang Judging.} 
Before judging, we firstly filter the offensive and dirty sentences. Then, to improve the annotation efficiency, we utilize Qwen2.5-72b to judge whether each utterance contains any slang terms. After that, we filter the sentences that contains repetitive slang terms and we obtain \textasciitilde776k out of 28M (2.8\%) sentences with the slang term. 

\noindent\textbf{Slang Extraction.} To ensure the data quality and avoid model bias, we separately utilize Qwen2.5-72b and Llama3.3-70b to extract the slang term. If both Qwen2.5-72b and Llama3.3-70b predict the same slang term and GPT-4o also approves that it is a slang term, we maintain such sentences (25k). 

\noindent\textbf{Explanation Generation.} With the advance Chinese ability of Qwen2.5-72b, we use it to generate Chinese explanation for each slang term. 

\noindent\textbf{Polysemy Annotation.} To fully investigate the impact of context on the slang term, we use GPT-4o to judge whether each slang term is polysemous. Consequently, we obtain 7,818 clear polysemous slang terms, and 17,258 non polysemous slang terms. 

Besides, for constructing the hard testset, we randomly sample 15\% instances (\emph{i.e.}, 1,165) from the polysemous slang terms. For the remaining 85\% (\emph{e.g.}, 6,653) polysemous slang terms, we use GPT-4o to generate possible meaning for each slang term. Finally, for each meaning, we further use GPT-4o to produce corresponding translation pairs that contains the same slang term but convey different sense. After this process, we obtain 26,400 sentence pairs for training.

\subsection{Annotation Quality Assessment}
\label{aqa}
To evaluate the quality of slang detection and cross-lingual slang explanation, we employ three annotators to judge whether the slang term is real and whether its explanation is correct targeting on the context over 200 randomly sampled data. Then, we measure the inter-annotator agreement. The inter-annotator agreements calculated by Fleiss’ kappa are 0.685 and 0.915 for slang detection and explanation, which indicates ``Substantial Agreement'' and ``Almost Perfect Agreement'' between three annotators, respectively. The level is consistent with previous work~\cite{liang-etal-2022-msctd} which can be considered as reliable.

\subsection{Dataset Statistics}
\label{dsa}
As shown in Table~\ref{Tbl:data}, the SlangDIT contains totally 44,823 English-Chinese utterance pairs with slang term, where each slang term has been annotated with explanation. According to slang terms, we split the dataset into 39,980 for training, 1,815 for validation, and 1,863 for testing\footnote{Note that there is no overlap of the slang term among training, validation and testing set.}. To keep roughly the same distribution of the utterance pair and avoid the model bias, we sample 39,980/1,815/1,863 utterance pairs from the original subtitles where both Qwen2.5-72b and Llama3.3-70b judge no slang terms for training/validation/testing, respectively. 

Based on the statistics in Table~\ref{Tbl:data}, the average numbers of tokens per utterance are about 7.8, 12.7, and 75.8 for English utterances (word level), Chinese translations (character level), and explanation (character level), respectively. 
\textbf{\begin{figure*}[t]
    \centering
    \includegraphics[width=0.99\textwidth]{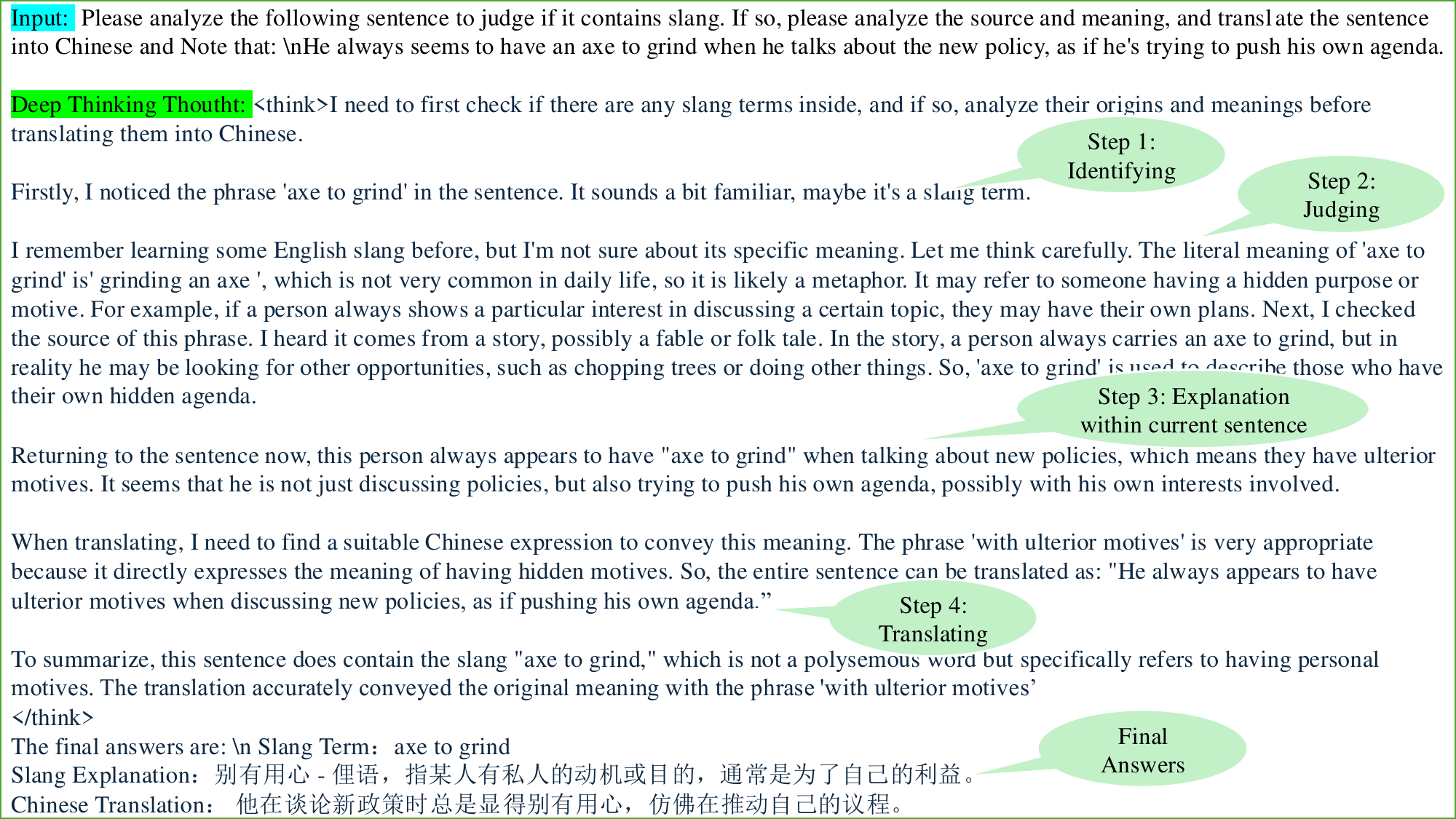}
    \caption{The generated deep thinking thought (training example) by DeepSeek-R1-Distill-Qwen-32B.
    }
    \label{fig.2}
\end{figure*}}

\subsection{Comparison with Related Datasets}
\label{rds}
The related datasets mainly involve three research fields: slang detection, cross-lingual slang explanation, and slang translation. 

In \textbf{slang detection}, there exist some dictionary-based dataset. For example, \citet{ni-wang-2017-learning} construct a Urban Dictionary, which is large-scale but the quality is poor. \citet{sun-etal-2022-semantically} manually annotate a small subset of 102 sentences from the Online Slang Dictionary (OSD). The Green’s Dictionary of Slang (GDoS)~\cite{adams2012green} cannot be publically distributed due to copyright restrictions. Besides, OpenSubtitles-Slang~\cite{sun-etal-2024-toward} only provides the slang definition without context. Most of these dataset only offer the slang term and its definition, which cannot be applied to SlangDIT due to the lack of polysemous labeling, diverse context and translation.

In \textbf{cross-lingual slang explanation}, ~\citet{noraset2017definition} propose the definition generation task. As time goes on, \citet{zhang-etal-2023-assisting} introduce cross-lingual definitions of (general) words in English, Chinese, and French using prompt learning. Recently, \citet{gluck2025clixcrosslingualexplanationsidiomatic} propose the task of cross-lingual explanations for idiomatic expressions.~\citet{sun-etal-2022-semantically} formulate the slang interpretation as the translation task, \emph{i.e.}, the slang term and its interpretation (both in English) are fed into a translation model.

In \textbf{slang translation}, closely related to our work is~\citet{sun-etal-2022-semantically} that show the slang interpretation in the same language can result in improved translation of slang in target language. In this work, they conduct experiment in English-German/French directions with a small-scale machine-translated dataset, which is not publicly available and not target on the study of polysemous slang terms.

The resources mentioned above are extensively used in corresponding fields of research and they even cover some sub-tasks in SlangDIT. However, our SlangDIT is different from them in terms of complexity and diversity. 

It is obvious that conducting three sub-tasks is more challenging due to the more complex scene. Furthermore, most of the above single dataset cannot be available and fails to conduct SlangDIT task. What's important, the slang term of above dataset is non-polysemous while ours is polysemous in different context which could be more difficult to interpret. Table~\ref{data:tr} provides information on the number of available modality, state of publicly accessible, and their constituent slang terms for all the datasets. Besides, compared with two existing slang dataset, SlangDIT's quantity of English is ten-times of the annotated utterances in most of dataset (except for Urban Dictionary, whose quality is limited). More importantly, the utterance of our SlangDIT comes from movie subtitles, which is natural and diverse than existing data. Particularly, SlangDIT provides an explanation in a cross-lingual language. 

\section{SlangOWL Model}

\textbf{Backbone.} We mainly utilize three LLMs as the backbones of our SlangOWL model: (1) Llama-3.1-8B-Instruct~\cite{dubey2024llama}; (2) Qwen2.5-7B-Instruct and (3) Qwen2.5-14B-Instruct~\cite{yang2024qwen2}.

\noindent\textbf{Thought Data.} To simulate cognitive process of humans in translation, we need the deep thinking thought samples to train our SlangOWL model. Given the promising reasoning ability of in existing o1 like models, we decide to use the advance DeepSeek-R1-Distill-Qwen-32B~\cite{deepseekai2025deepseekr1incentivizingreasoningcapability} model. Specifically, we provide four key elements: slang term, polysemy, cross-lingual explanation and translation to the DeepSeek-R1-Distill-Qwen-32B model and prompt it to generate the deep thinking thought for each training instance. The reasoning process roughly includes four steps: 1) identifies whether the sentence contains a slang, and then 2) judges whether the slang is polysemous and analyze its possible meaning. Further, 3) provides the best cross-lingual explanation of the slang term targeting on the current context. Finally, 4) according to the thought, the model offers a suitable translation. We list the generated thought example in Figure~\ref{fig.2} and list the prompt in Figure~\ref{fig.3} of Appendix.

\begin{table*}[t]
\centering
\scalebox{0.8}{
\setlength{\tabcolsep}{0.5mm}{
{
\begin{tabular}{llccccccccccc}
\toprule[1pt]
\multicolumn{1}{c}{\multirow{2}{*}{\textbf{Types}}}&\multicolumn{1}{c}{\multirow{2}{*}{\textbf{Models}}} & \multicolumn{3}{c}{\textbf{Slang Detection}} & \multicolumn{3}{c}{\textbf{XLSE}}    & \multicolumn{5}{c}{\textbf{Translation}}                    \\
\cmidrule(r){3-5} \cmidrule(r){6-8} \cmidrule(r){9-13} \multicolumn{1}{c}{}  &    & P   & R  & F1  & R-1   & R-2 & R-L & BLEU  & Comet & GRB & CometK & GRF            \\ \midrule[1pt]
\multirow{11}{*}{\emph{Vanilla LLMs}} 
&Llama-3.1-8B-Instruct   & 9.44 & 12.96 & 10.61& 17.43 & 2.78 & 11.79 &12.69 & 65.17 &54.24 &  59.96 & 70.02 \\
&Qwen2.5-7B-Instruct   & 74.72 & 52.04 & 56.94 & 25.00 & 7.21 & 17.82 & 16.98 &72.41 & 66.54&70.52 & 72.66    \\
&Qwen2.5-14B-Instruct   &66.77 & 75.60 &70.46 & 35.48&10.31 &25.63& 17.66 & 73.82 & 68.57& 72.36 & 75.58 \\\cdashline{2-13}[4pt/2pt]
&DRT-o1-8B    &0.0 & 0.0 & 0.0 & 5.00 & 3.49 & 3.15 &12.43&39.21&54.10&39.33&58.24\\
&DRT-o1-7B   &4.72 & 8.94 & 6.12 & 15.88&1.13 &13.01 &14.94 &67.62&50.31&66.47&69.54\\
&DRT-o1-14B    &63.95 & 44.31 & 50.71 &24.75 & 5.82 & 17.04 & 18.45 &70.25&65.24&68.10&73.89\\
&DS-R1-D-Llama-8B   &61.57&66.12 & 63.28 &27.33 & 5.13 & 19.04 &15.76 &71.52&58.01&70.82 & 64.59\\
&DS-R1-D-Qwen-7B   &33.67 & 43.06 & 37.79 &22.21&4.13&15.55 & 12.89 &64.61&49.17 &61.65 &42.05\\
&DS-R1-D-Qwen-14B   &68.41 & 71.41 & 69.35 &30.85&7.16&21.82 & 18.46 & 73.18 &64.77 & \textbf{72.56} & 73.37\\
&DS-R1-D-Qwen-32B   &60.90 & 63.33 & 61.19 &33.04&8.28&23.21 & 19.12&73.69&63.40&{72.47}&73.39\\

&QwQ-32B-preview   &38.92 & 55.45 & 45.42 &27.07&6.34 & 19.63 & 14.07&72.87&67.98&70.72&\underline{76.84}\\

\midrule
\multirow{3}{*}{\emph{SFT w/o cot}} 
&SFT-8B    &\underline{85.84} &61.22 & 67.03 &28.88&17.85&25.48 & 22.91  &73.17 &66.39 & \underline{72.50} &71.35 \\ 
&SFT-7B  &84.19 & 70.67 & 74.74 & 28.93 &17.87 & 25.53 &23.65& 73.70 & 66.98& 71.12  & 72.71 \\
&SFT-14B   &\textbf{86.74} & 74.15 & 78.21 & 29.98 & 18.78&26.26 &\underline{24.51}& 73.62 & \underline{69.91} & 71.29 & {76.00}  \\ 
\midrule
\multirow{3}{*}{\emph{Deep Thinking}} 
&SlangOWL-8B  & {84.35} & \underline{87.30}$^{\dagger\dagger}$ & \underline{85.33}$^{\dagger\dagger}$ &\underline{54.77}$^{\dagger\dagger}$ & \underline{34.16}$^{\dagger\dagger}$ & \underline{51.28}$^{\dagger\dagger}$ &23.47$^{\dagger}$& 73.89$^{\dagger}$ & {68.40}$^{\dagger\dagger}$& 70.72 &{75.08}$^{\dagger\dagger}$\\ 
&SlangOWL-7B  & {82.59} & {86.49}$^{\dagger\dagger}$ & {84.02}$^{\dagger\dagger}$  & {58.50}$^{\dagger\dagger}$ & {33.16}$^{\dagger\dagger}$ & {52.51}$^{\dagger\dagger}$ & 24.10$^{\dagger}$ & \underline{74.04} & {68.38}$^{\dagger\dagger}$& {71.60}$^{\dagger}$ &{74.64}$^{\dagger\dagger}$ \\
&SlangOWL-14B   & {85.17} &\textbf{88.78}$^{\dagger\dagger}$ & \textbf{86.47}$^{\dagger\dagger}$ &\textbf{59.85}$^{\dagger\dagger}$& \textbf{34.31}$^{\dagger\dagger}$ &\textbf{53.72}$^{\dagger\dagger}$  &\textbf{24.94} & \textbf{74.20}$^{\dagger}$ & \textbf{71.52}$^{\dagger\dagger}$&  {71.38} & \textbf{77.52}$^{\dagger\dagger}$  \\ \bottomrule[1pt]
\end{tabular}
}}}
\caption{Experimental results (\%) on general test set. `XLSE' denotes cross-lingual slang explanation. `DS-R1-D' denotes `DeepSeek-R1-Distill'. ``$^{\dagger}$'' and ``$^{\dagger\dagger}$'' denote that statistically significant better than the best result of the counterpart (\emph{e.g.}, SlangOWL-14B vs. SFT-14B) with t-test {\em p} \textless \ 0.05 and {\em p} \textless \ 0.01 hereinafter, respectively. The best and second best results are \textbf{bold} and \underline{underlined}, respectively.} 
\label{table:main_res1}
\end{table*}

\section{Experiments}

\subsection{Experimental Setups}

\noindent\textbf{Comparison Models.}
We include three types of baselines: 1) Vanilla instructed models; 2) Vanilla reasoning models; and 3) supervised fine-tuned models without chain-of-thought (SFT w/o cot). Please refer to Appendix~\ref{app_model} for details. For more training details and inference details, please refer to Appendix~\ref{app_details}.

\noindent\textbf{Metrics.}
For \emph{slang detection}, we adopt P, R, and F1 as the metric following previous work~\cite{sun-etal-2024-toward}. For \emph{cross-lingual slang explanation}, we use ROUGE-1, ROUGE-2, ROUGE-L, where ROUGE-L evaluate n-grams overlap between the generated explanation and corresponding references~\cite{gluck2025clixcrosslingualexplanationsidiomatic}. We refer ROUGE-* as R-* in the Table for simplicity. For \emph{translation}, we report reference-based and reference-free scores to evaluate the model translations. In reference-based metrics, we use BLEU~\cite{papineni-etal-2002-bleu}, and Comet~\cite{rei-etal-2020-comet}. In reference-free metrics, we use CometKiwi (refered as CometK~\cite{rei-etal-2022-cometkiwi}). Here, BLEU is to evaluate n-grams overlap between the generated explanation and corresponding references, and CometScore evaluate the semantic similarity of translations against references. The CometK uses a language model to judge whether a translation conveys the semantics of the source sentence. Besides, following recent studies~\cite{kocmi2023large,wang-etal-2023-chatgpt,wang2024drto1optimizeddeepreasoning} that show the strong ability of LLMs in NLP evaluation, we also use GPT-4o as evaluators in reference-based and reference-free manner, which we refer to as \emph{GRB} and \emph{GRF}, respectively. For calculation details and evaluation prompts, please refer to Appendix~\ref{sec:gpt4o_eval}.




\begin{table*}[t]
\centering
\scalebox{0.85}{
\setlength{\tabcolsep}{0.4mm}{
{
\begin{tabular}{lccccccccccc}
\toprule[1pt]
\multicolumn{1}{c}{\multirow{2}{*}{\textbf{Models}}} & \multicolumn{3}{c}{\textbf{Slang Detection}} & \multicolumn{3}{c}{\textbf{XLSE}}    & \multicolumn{5}{c}{\textbf{Translation}}                    \\
\cmidrule(r){2-4} \cmidrule(r){5-7} \cmidrule(r){8-12} \multicolumn{1}{c}{}                       & P   & R  & F1              & R-1   & R-2 & R-L & BLEU  & Comet & GRB & CometK & GRF            \\ \midrule[1pt]
Qwen2.5-14B-Instruct   & 49.61 &55.36 &51.39 & 39.01&10.66&27.76 &12.44  & 67.14 &67.19 & 63.67 &75.72 \\\cdashline{1-12}[4pt/2pt]
DRT-o1-14B    &39.25 & 40.86 & 39.65 &36.98&8.71&25.61 &13.76 & 57.75 &68.32 &54.02 &79.16\\
DeepSeek-R1-Distill-Qwen-14B   &56.88 & 64.21 & 59.15 &38.73&9.31&27.90 & 14.02 &65.42 &63.42 &63.84 &73.28\\
DeepSeek-R1-Distill-Qwen-32B   &34.62 & 45.01 & 38.02 &39.57&10.37&28.33 & 14.91 &64.37&63.51&64.02 &73.25\\
QwQ-32B-preview   &18.44 & 19.55 & 18.75 &32.55&7.65&23.77 &10.26 &66.72 &62.19 &62.44 &71.92\\
\midrule
SFT-14B    & \underline{78.09} & \underline{83.94} & \underline{80.05} &\underline{57.68} &\underline{36.99} & \underline{50.66} &\underline{19.18} & \underline{68.26} &\underline{69.62} & \textbf{64.56}  &\underline{79.64}  \\ 
\midrule
SlangOWL-14B   &\textbf{88.86}$^{\dagger\dagger}$  &\textbf{91.50}$^{\dagger\dagger}$  &\textbf{89.60}$^{\dagger\dagger}$ &\textbf{60.39}$^{\dagger\dagger}$ & \textbf{39.18}$^{\dagger\dagger}$ & \textbf{53.55}$^{\dagger\dagger}$ & \textbf{21.41}$^{\dagger\dagger}$  & \textbf{68.49} &\textbf{70.02} & \underline{64.49} &\textbf{82.35}$^{\dagger\dagger}$   \\ \bottomrule[1pt]
\end{tabular}
}}}
\caption{Experimental results (\%) on the hard test set.} 
\label{table:main_res2}
\end{table*}




\subsection{Main Results}
Table~\ref{table:main_res1} shows the main results on the general testset where each instance contains either a non-polysemous slang term or no slang term. Table~\ref{table:main_res2} presents additional results on the hard testset where each instance contains a polysemous slang term.

\subsubsection{Results on the General Testset}

\paragraph{Results on Slang Detection.}
\label{ssec:ende}
Under the \emph{vanilla} setting in Table~\ref{table:main_res1}, the reasoning models significantly surpasses the instructed models with the same model scale in terms of F1 score. The bigger models also show better performance than smaller ones, proving that larger models owns better memory capacity. There is no doubt that the \emph{SFT w/o cot} consistently outperforms their counterpart, \emph{e.g.}, the SFT-14B beats Qwen2.5-14B-Instruct, DeepSeek-R1-Distill-Qwen-14B and DRT-o1-14B, even showing better performance than QwQ-32B-preview and DeepSeek-R1-Distill-Qwen-32B. However, the best F1 score (78.21\% with SFT-14B) of these models is still substantially lower than the series model of SlangOWL. It shows that the slang term is not just a simple memory task and needs deep thinking to judge whether it is a real slang term. Although o1-like models can conduct reasoning, they fail to identify the correct slang terms. Our proposed SlangOWL models have deep reasoning ability, which analyze the novel and possible slang terms, their background and sense. Therefore, SlangOWL models set a state-of-the-art F1 score (86.47\%). 


\paragraph{Results on Cross-lingual Slang Explanation.}
\label{ssec:chen}
Although the instructed and reasoning models achieved good F1 score on slang detection, they cannot generate good slang explanation in Chinese. As shown in Table~\ref{table:main_res1}, the ROUGE scores are much lower than supervised fine-tuned models, which shows that they only `\emph{know which phrase is slang term, but not know why it is}'. Furthermore, the ROUGE scores of \emph{SFT w/o cot} are remarkably worse than SlangOWL models. This demonstrates that the \emph{SFT w/o cot} also cannot truly master the meaning of the slang term since the real sense of slang term always goes beyond its original meaning and shows extended meaning. In contrast, slangOWL models consistently outperform the comparison methods, achieving significantly better ROUGE scores. It shows that the proposed deep thinking model not only know its original meaning but also get its deeper implications. Therefore, the proposed slangOWL models offer correct cross-lingual slang explanation.

\paragraph{Results on Translation.}
In terms of reference-based scores (\emph{i.e.}, BLEU, ComentScore, and GRB), interestingly, although the vanilla models fails to achieve good results of slang detection and cross-lingual slang explanation, some models still obtain good translation results (\emph{e.g.}, Qwen2.5-14B-Instruct and QwQ-32B-preview). Obviously, the \emph{SFT w/o cot} and SlangOWL consistently surpass their counterparts with the help of good results of slang detection and cross-lingual slang explanation. Armed with the deep thinking, the SlangOWL obtains the highest scores.

In terms of reference-free scores (\emph{i.e.}, CometKivi and GRF), we can observe similar findings on reference-based scores. However, we find that the DS-R1-D-Qwen-14B model, a deep reasoning model, achieves the best results in CometK score while greatly underperforms in other metrics. The reason may be that this model generates some words that highly fitting the source words. Except that, the SlangOWL consistently outperforms all previous models once again (including QwQ-32B-preview and DeepSeek-R1-Distill-Qwen-32B), showing its superior performance. 

\paragraph{Overall Results.}
Overall, with better results of slang detection and cross-lingual slang explanation, the model can achieve better translation results (SlangOWL vs. SFT w/o cot; SFT-based vs. vanilla). It shows that correct understanding of slang terms plays a key role in translate sentence with slang terms. What's important, \emph{the ability of slangDIT indeed can decides whether the LLMs can go beyond superficial meaning of slang term} and thus prove the value of slangDIT benchmark.

\subsubsection{Results on the Hard Testset}
Table~\ref{table:main_res2} shows the results on the hard testset where each instance includes a polysemous slang term. We can find that the instructed model and reasoning model performs worse in terms of all three tasks, showing that they all struggle to judge whether it is a slang term, and have a difficulty in understanding its real sense targeting on the current context and thus leading to unsatisfactory translation results. Meanwhile, we observe that different learning manners of simple fine-tuning and deep thinking reflect great difference on effects. It shows that there is much room for further improvement using other more advanced learning methods.

Compared with the results in Table~\ref{table:main_res2}, we find that our SlangOWL-14B performs much better on the hard testset (89.6\% vs. 86.47\%), which shows that it has higher ability to correctly understanding the polysemous slang term in different contexts and thus translate the sentence well. 

\begin{table}
    \centering
	\begin{minipage}{0.49\linewidth}
		\centering
\newcommand{\tabincell}[2]{\begin{tabular}{@{}#1@{}}#2\end{tabular}}
\scalebox{0.82}{
\setlength{\tabcolsep}{0.7mm}{
\begin{tabular}{rrr}
\toprule
$\textbf{SFT-14b}$& \multicolumn{1}{c}{GoodT} & \multicolumn{1}{c}{BadT} \\
\midrule
\multirow{1}{*}{\tabincell{c}{CSU: 76.31}}
 &75.34     &24.66         \\
\hline
\multirow{1}{*}{\tabincell{c}{WSU: 23.69}}
&27.54 &72.46       \\
\bottomrule
\end{tabular}}}
	\end{minipage}
	\hfill
	\begin{minipage}{0.49\linewidth}
		\centering
\newcommand{\tabincell}[2]{\begin{tabular}{@{}#1@{}}#2\end{tabular}}
\scalebox{0.82}{
\setlength{\tabcolsep}{0.7mm}{
\begin{tabular}{rrr}
\toprule
$\textbf{SlangOWL}$& \multicolumn{1}{c}{GoodT} & \multicolumn{1}{c}{BadT} \\
\midrule
\multirow{1}{*}{\tabincell{c}{CSU: 82.15}}
 &88.41     &11.59         \\
\hline
\multirow{1}{*}{\tabincell{c}{WSU: 17.85}}
&18.63  &81.37      \\
\bottomrule
\end{tabular}}}
	    \end{minipage}
\caption{Results (\%) of investigation whether the correct slang understanding helps on the hard testset.}
\label{tbl:w_f}
\end{table}


\subsection{Analysis}
\noindent\textbf{Is the Correct Slang Understanding helpful to Translation?}
Before investigating whether the correct slang understanding works, we define some metrics: 1) the correct slang understanding means the model not only correctly predict the slang term and the ROUGE-L score is greater than 0.4, we denote it as CSU, otherwise we denote is as WSU (wrong slang understanding); 2) good translation means both the GRB and GRF are greater than 70 and 80, respectively, we denote it as GoodT, otherwise we denote it BadT (bad translation). Based on the definition, we calculate these metric for SFT-14B and SlangOWL-14B models. 

The results are shown in Table~\ref{tbl:w_f}. We observe that in CSU, the GoodT score is significantly better than BadT score with both SFT-14B and SlangOWL-14B models while under WSU, the BadT score significantly wins. It shows that the correct slang understanding indeed helps for better translation and also reflects that the deep thinking has a positive impact on the SlangDIT task.

\noindent\textbf{Compared with Models Translating only}.
Since the vanilla models do not optimized for slang understanding, they performs worse on slang detection and explanation that further result in bad translation. In this section, we prompt these vanilla models for translation only to protect them from suffering understanding the slang term. Besides, we also train a model with translation pair only based on Qwen2.5-14B-Instruct, denoted as SFT-Trans-14B.

The results are listed in Table~\ref{translationly}. We conclude the following findings: 1) The vanilla models (including DeepSeek-R1-Distill-Qwen-32 and QwQ-32B-preview models) performs translation worse on the sentence with a polysemous slang term, showing that the ability of SlangDIT indeed decides whether the LLMs can go beyond superficial meaning. 2) The SlangOWL-14B significantly outperforms the SFT-Trans-14B model, showing that our SlangOWL model have the ability of deep thinking, \emph{i.e.}, first identifying the slang term, then understanding its background and usage targeting on the current context, and finally providing suitable and satisfactory translations.

\noindent\textbf{Case Study}.
We present one case study in Appendix~\ref{app_cs} to intuitively show how the deep thinking helps to translate polysemous slang terms well.


\section{Related Work}
We have introduce some task-related work in~\autoref{rds} including slang detection, cross-lingual slang explanation, and translation. Next, we present some work in reasoning.

With the emergency of OpenAI O1~\cite{openai_o1_2024} model, some studies have been devoted to the reasoning tasks (\emph{e.g.}, math and coding)~\cite{zhang2024o1,huang2024o1,qin2024o1,deepseekai2025deepseekr1incentivizingreasoningcapability}. In the context of translation, ~\citet{zhao2024marco} proposes Marco-o1 for open-ended text generation and show the potentiality of the long thought reasoning for translation. More recently, \cite{wang2024drto1optimizeddeepreasoning} introduces long Chain-of-Thought for literature translation and achieves good results. Different from them, we mainly focus on benchmarking interpretative slang translation task, which is more complex since we need conduct three tasks jointly and ensure the first results are valid to the translation. Besides, we propose a deep thinking model according to the task characteristic of SlangDIT.

\begin{table}[t]
\centering
\newcommand{\tabincell}[2]{\begin{tabular}{@{}#1@{}}#2\end{tabular}}
\scalebox{0.75}{
\begin{tabular}{lccccc}
\toprule
{Models} & BLEU / {Comet}  / {GRB} / {CometK}  / {GRF}\\
\midrule
Qwen2.5-14B-Instruct            &13.54  / 67.52  / \underline{71.05} / 66.67 / \underline{79.68}\\
QwQ-32B-preview          &10.91 / 65.85 / 62.20 / 66.52 / 71.92 \\
DS-R1-D-Qwen-32B        &16.42 / \underline{67.89} / 70.23  / \underline{67.62} / 78.73\\
DS-R1-D-Qwen-14B  &15.04 / 66.26 / 65.95  / 67.17 / 74.62\\
DRT-o1-14B  &11.83 / 67.02 / 68.32 / \textbf{68.13} / 79.17 \\\cdashline{1-4}[4pt/2pt]
SFT-Trans-14B &\underline{19.91} / 67.50 / \textbf{71.38} / 62.75 / 76.70\\\cdashline{1-4}[4pt/2pt]
SlangOWL-14B& \textbf{21.41} / \textbf{68.49} / {70.02} / {64.49} / \textbf{82.35} \\
\bottomrule
\end{tabular}}
\caption{Translation results on the hard testset. }
\label{translationly} 
\end{table}

\section{Conclusion}
In this paper, we introduce a new interpretative slang translation task that consists of three sub-tasks: slang detection, cross-lingual slang explanation, and translation. Then, we construct a interpretative slang translation dataset named SlangDIT. Finally, we propose a deep thinking model named SlangOWL and demonstrate the importance of slang detection and explanation for SlangDIT task. 

\section*{Limitation}
While we introduce a SlangDIT dataset and propose a deep thinking model named SlangOWL, there are some limitations worth considering to study in future work: (1) In this study, we only provide the slang term in English, and future work could extend our dataset to more language pairs, \emph{e.g.}, English to French, Chinese to English; (2) This work does not conduct experiments on more large models due to limited resources, where future work could verify our method on larger models; (3) This work does not conduct pipline experiments that firstly optimize for slang detection and then offer the results to translation model since our work mainly focuses on introducing a new SlangDIT task that simultaneously conducts three subtasks. In this process, we hope the correct understanding of slang term can help translation.

\section*{Ethical Considerations}
\label{ec}
In this section, we discuss the main ethical considerations of SlangDIT: (1) Intellectual property protection. The English utterance of SlangDIT is from MSCTD dataset~\cite{liang-etal-2022-msctd}. For our slang terms and cross-lingual explanations, its permissions are granted to copy, distribute and modify the contents under the terms of the \href{https://en.wikipedia.org/wiki/Wikipedia:Text_of_Creative_Commons_Attribution-ShareAlike_3.0_Unported_License}{Creative Commons AttributionShareAlike 3.0 Unported License} and \href{https://www.wikidata.org/wiki/Wikidata:Text_of_the_Creative_Commons_Public_Domain_Dedication}{Creative Commons CC0 License}, respectively. (2) Privacy. The data source are publicly available movies. Its collection and slang/explanation annotation procedure is designed for interpretative slang translation purpose, and does not involve privacy issues. (3) Compensation. During the slang or explanation annotation, we use publicly available Qwen2.5-72b and Llama3.3-70b models. For polysemy annotation, we use GPT-4o and we have paid for them according to the official price. (4) Potential problems. While principled measures are taken to ensure the quality of the dataset, there might still be potential problems with the dataset quality due to the uncontrollability of models, which may lead to incorrect translations in applications. However, moderate noise is common in large-scale modern translators, even for human translated sentences, which should not cause serious issues.


\bibliography{custom}

\clearpage
\newpage
\appendix

\section{Backbones and Other Details}
\label{app_details}
\noindent \textbf{Backbones.}
We mainly utilize the following three LLMs as the backbones for SlangDIT task: (1) Llama-3.1-8B-Instruct~\cite{dubey2024llama}\footnote{\url{https://huggingface.co/meta-llama/Llama-3.1-8B-Instruct}}; (2) Qwen2.5-7B-Instruct\footnote{\url{https://huggingface.co/Qwen/Qwen2.5-7B-Instruct}} and (3) Qwen2.5-14B-Instruct~\cite{yang2024qwen2}\footnote{\url{https://huggingface.co/Qwen/Qwen2.5-14B-Instruct}}.

\noindent \textbf{Implementation Details.}
During training, Llama-Factory~\cite{zheng-etal-2024-llamafactory} is used to instruct-tune LLMs. Following~\citet{wang2024drto1optimizeddeepreasoning}, all LLMs are tuned on two 8$\times$NVIDIA A100 GPUs (40G) with 1e-5 learning rate. We set gradient accumulation to 16 and batch size to 1, which gives us 2*8*16*1 batch in total.
We use the DeepSpeed optimization~\cite{rasley2020deepspeed}, and set ZeRO-3 optimization.
Following~\citet{qin2024o1}, we set the number of training epochs to 3, and the training process costs about 48, 43 and 90 GPU hours for 8b, 7B and 14B models, respectively.

\noindent \textbf{Inference Details.}
During inference, we use vLLM toolkit~\cite{kwon2023efficient}\footnote{\url{https://github.com/vllm-project/vllm}} to accelerate the model generation for all models. We use the sampling decoding strategy with 0.1 temperature, and set the repetition penalty to 1.05.

\section{Comparison Models}
\label{app_model}
We include three types of baselines: 1) Vanilla instructed models; 2) Vanilla reasoning models\footnote{During inference of vanilla instructed models and vanilla reasoning models, we prompt them to directly conduct three tasks and use two-shot prompting to enhance their performance.}; and 3) supervised fine-tuned models. Please refer to Appendix~\ref{app_model} for details.

\noindent\textbf{Vanilla Instructed Models}. 
We use three backbones as the comparison model: Llama-3.1-8B-Instruct, Qwen2.5-7B-Instruct and Qwen2.5-14B-Instruct.

\noindent\textbf{Vanilla Reasoning Models}. 
Recently, o1-like models have achieved significant results on reasoning tasks. Therefore, we include some models to compare with our deep thinking method. These models are QwQ-32B-preview~\cite{qwq-32b-preview}, DeepSeek-R1-Distill-Llama-8B, DeepSeek-R1-Distill-Qwen-7B, DeepSeek-R1-Distill-Qwen-14B, DeepSeek-R1-Distill-Qwen-32B~\cite{deepseekai2025deepseekr1incentivizingreasoningcapability}, DRT-o1-7B, DRT-o1-8B
and DRT-o1-14B~\cite{wang2024drto1optimizeddeepreasoning}. We refer `DeepSeek-R1-Distill-Qwen' as `DeepSeek-R1-D-Llama' in Table~\ref{table:main_res1}.

\noindent\textbf{Supervised Fine-tuned Models without Chain-of-Thought (SFT \emph{w/o} cot)}.  
For a fair comparison, we train three models based on Llama-3.1-8B-Instruct, Qwen2.5-7B-Instruct and Qwen2.5-14B-Instruct with the same training data as our SlangOWL model without deep thinking process (denoted as SFT w/o cot).

\section{Details of Metric Calculation and GPT-4o Evaluator}
\label{sec:gpt4o_eval}
We use the \textit{sacrebleu} toolkit\footnote{\url{https://github.com/mjpost/sacrebleu}} to calculate the corpus-level BLEU. To calculate Comet and CometK, we leverage the official codes\footnote{\url{https://github.com/Unbabel/COMET}} and the official models\footnote{\url{https://huggingface.co/Unbabel/wmt22-cometkiwi-da} and \url{https://huggingface.co/Unbabel/wmt22-comet-da}}. For calculating GRB and GRF, we randomly select 400 samples from the (hard) testing set since they need API costs. The prompts of reference-based (GRB) and reference-free (GRF) metric are listed in Figure~\ref{fig.prompts}. Both prompts are borrow from~\citet{kocmi2023large} with some adaptions to slang translation scene.

\textbf{\begin{figure*}[t]
    \centering
    \includegraphics[width=0.99\textwidth]{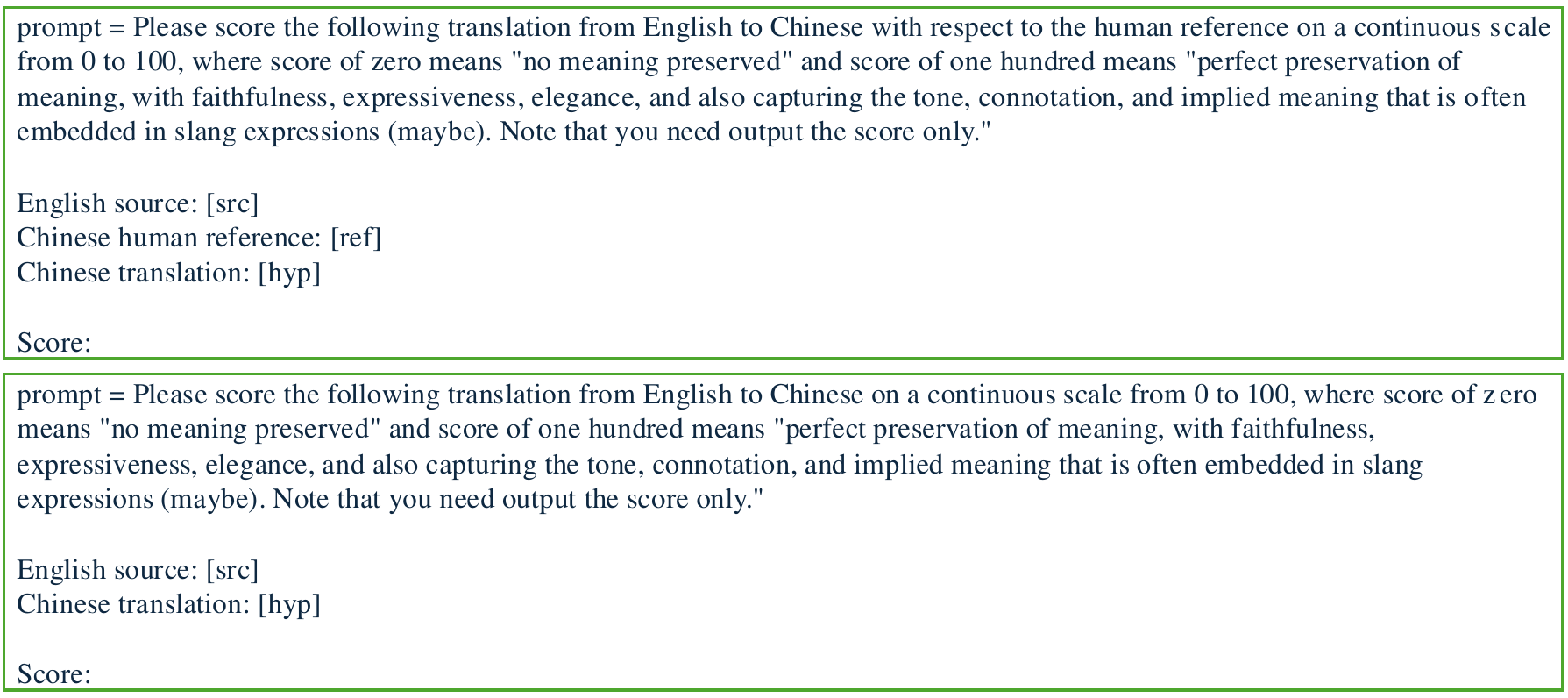}
    \caption{Two prompts used in GRB and GRF during evaluation via GPT-4o where ``[src]'',``[ref]'' and ``[hyp]'' mean the source sentence, human translation and model translation, respectively.
    }
    \label{fig.prompts}
\end{figure*}}

\textbf{\begin{figure*}[t]
    \centering
    \includegraphics[width=0.99\textwidth]{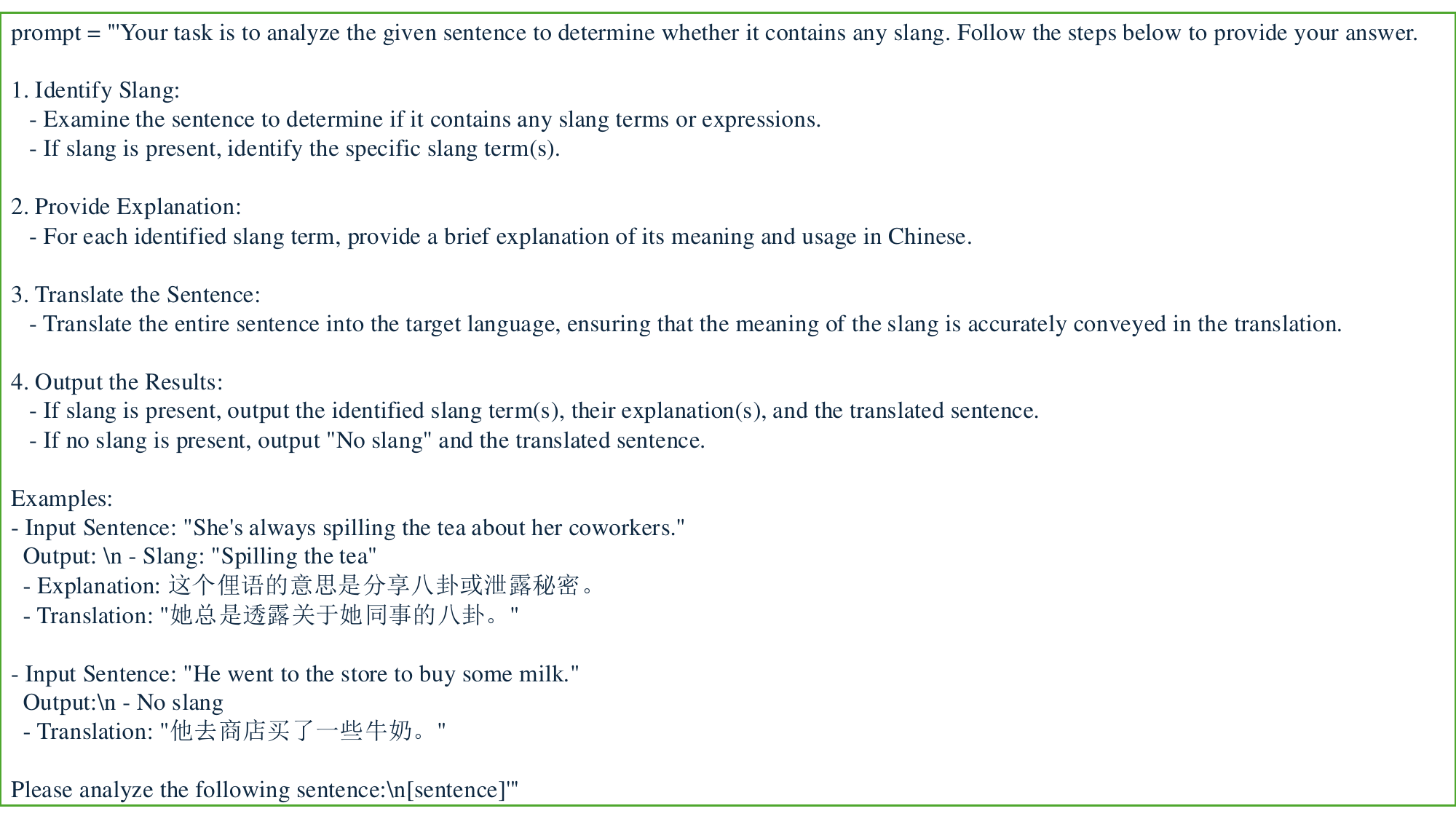}
    \caption{The prompt used in prompting vanilla models where ``[sentence]'' means the source sentence.
    }
    \label{fig.prompts_vanilla}
\end{figure*}}

\section{Prompts used in Prompting Vanilla Models}

When prompting vanilla instructed models and vanilla reasoning models in Table~\ref{table:main_res1} and Table~\ref{table:main_res2}, we use the prompt in Figure~\ref{fig.prompts_vanilla} to ask these models to generate their answers following the format. 

During prompting vanilla instructed models and vanilla reasoning models in Table~\ref{translationly}, we ues the prompt: `Translate the sentence into Chinese and output only the translation:\\n[sentence]'. Note that for DRT-o1-14B model, we strictly follow their prompts in the official repository\footnote{\url{https://github.com/krystalan/DRT-o1}}.

\section{Case Study}
\label{app_cs}
We list translation results of the models in Table~\ref{translationly}, where vanilla models are not struggle to the understanding of slang term. 

In the case, `Annie Oakley' is polysemous phrase which denotes a name or `slang term'. In all contrast models, they all take it as a name during translation. Although the QwQ-32B-preview realizes that it may convey extended meaning, it fails to and only translates it as `playing the role of Annie Oakley'. However, if people do not know who Annie Oakley is, it is hard for them to understand such translation. That is, such translation still not convey intended meaning the speaker said. In contrast, the SlangOWL-14B model can convey the ideas well, showing the effectiveness of the proposed model which can list its thought step-by-step.

\section{Compared to Stronger Models}
In this section, we compared with some stronger models (\emph{e.g.}, commercial system: Google Translator and other much advanced large-scale LLM models: Llama3.3-70B, Qwen2.5-72B, and GPT-4o). The results are shown in~\ref{stronger}, which demonstrate the effectiveness of the proposed method.

\begin{table}[t]
\centering
\newcommand{\tabincell}[2]{\begin{tabular}{@{}#1@{}}#2\end{tabular}}
\scalebox{0.75}{
\begin{tabular}{lccccc}
\toprule
{Models} & BLEU / {Comet}  / {GRB} / {CometK}  / {GRF}\\
\midrule
Google	&21.08	/ 67.93/	70.97/	67.53/	77.21\\
Llama3.3-70B            &18.23	/ 65.14	/ 68.78 /	65.21	/ 78.45\\
Qwen2.5-70B          &20.12	/ 67.24	/ 69.47	/ 67.51	/ 79.24 \\
GPT-4o	        &21.08	/ \textbf{68.93}	/ \textbf{70.58}	/ \textbf{69.58}	/ \textbf{82.78}\\
SlangOWL-14B& \textbf{21.41} / {68.49} / {70.02} / {64.49} / {82.35} \\
\bottomrule
\end{tabular}}
\caption{Compared to stronger models on the hard testset. }
\label{stronger} 
\end{table}

\textbf{\begin{figure*}[t]
    \centering
    \includegraphics[width=0.99\textwidth]{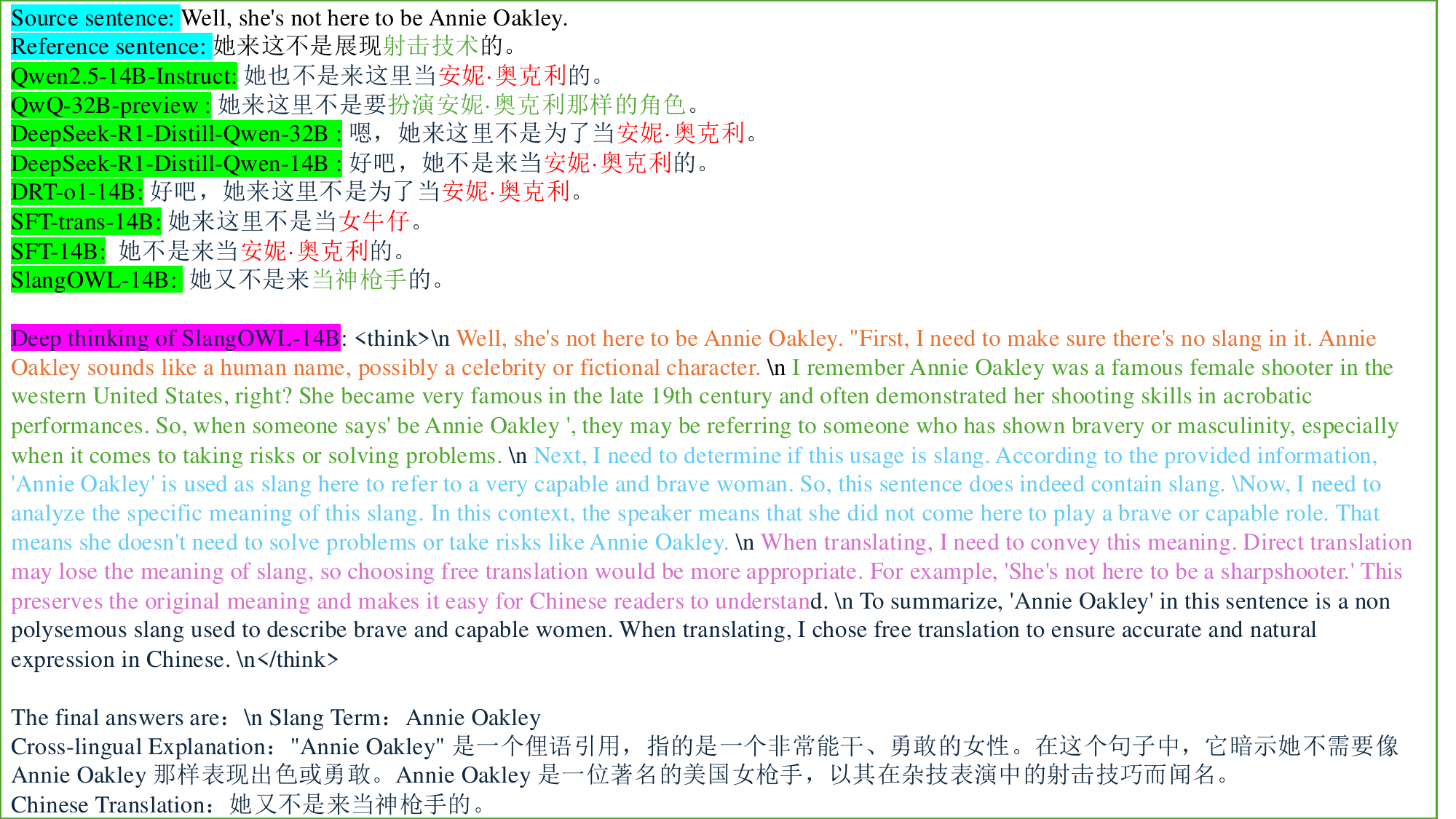}
    \caption{Case Study.
    }
    \label{fig.case_study}
\end{figure*}}

\textbf{\begin{figure*}[t]
    \centering
    \includegraphics[width=0.99\textwidth]{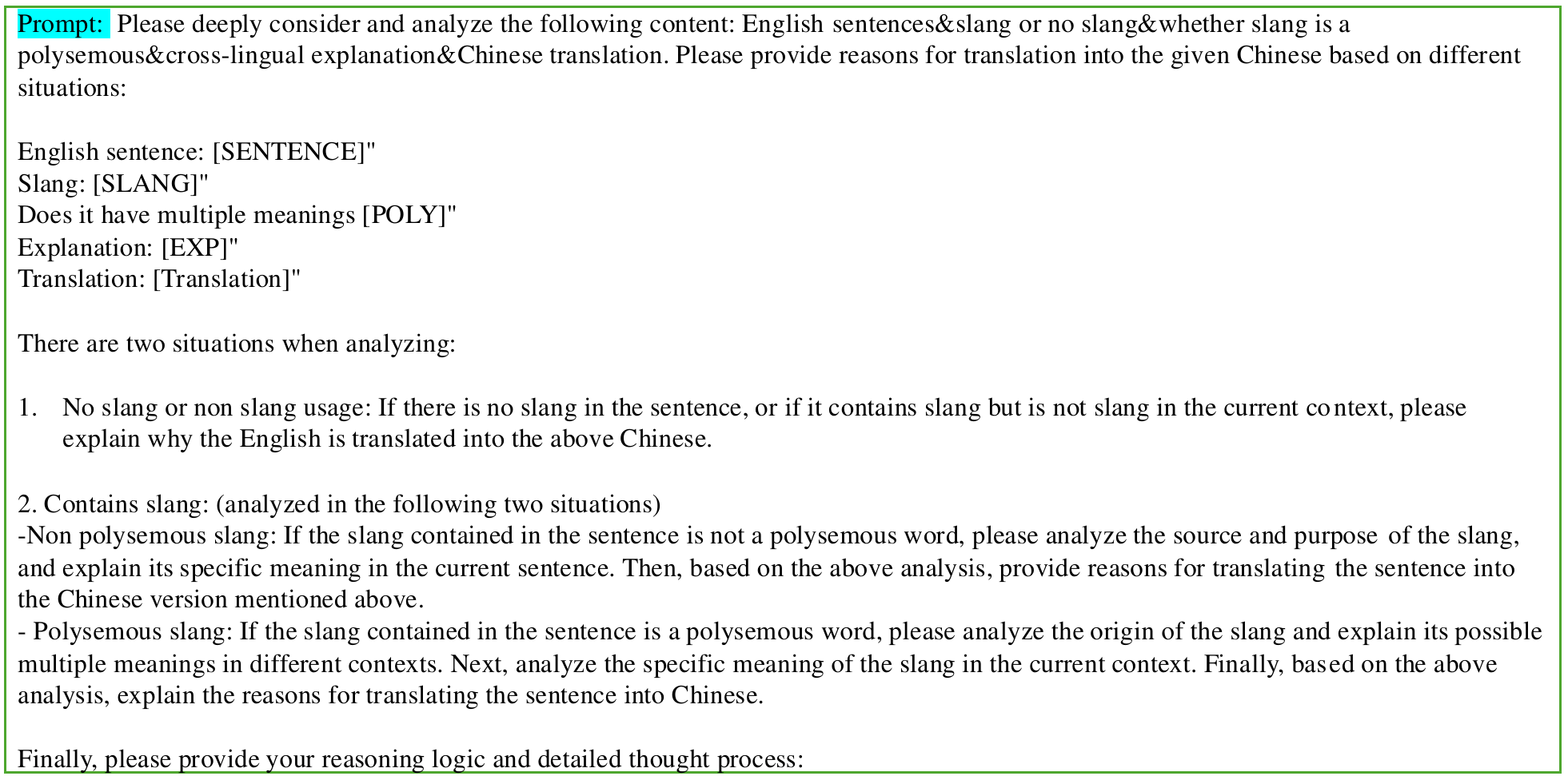}
    \caption{The prompt used in generating deep thinking thought by DeepSeek-R1-Distill-Qwen-32B.
    }
    \label{fig.3}
\end{figure*}}

\textbf{\begin{figure*}[t]
    \centering
    \includegraphics[width=0.99\textwidth]{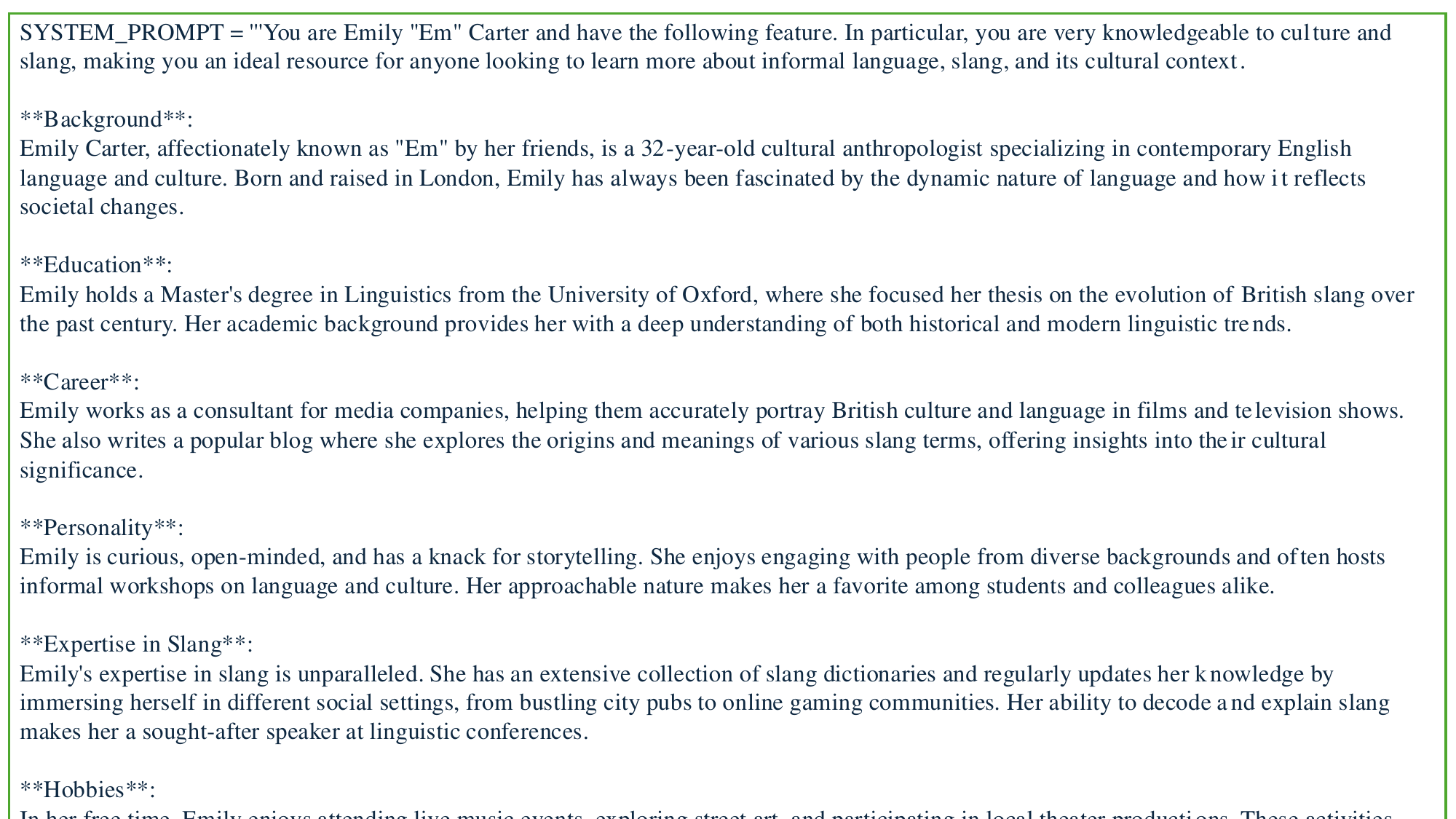}
    \caption{The system prompt used in the section of Annotation Procedure.}
    \label{annotaion_app}
\end{figure*}}

\textbf{\begin{figure*}[t]
    \centering
    \includegraphics[width=0.99\textwidth]{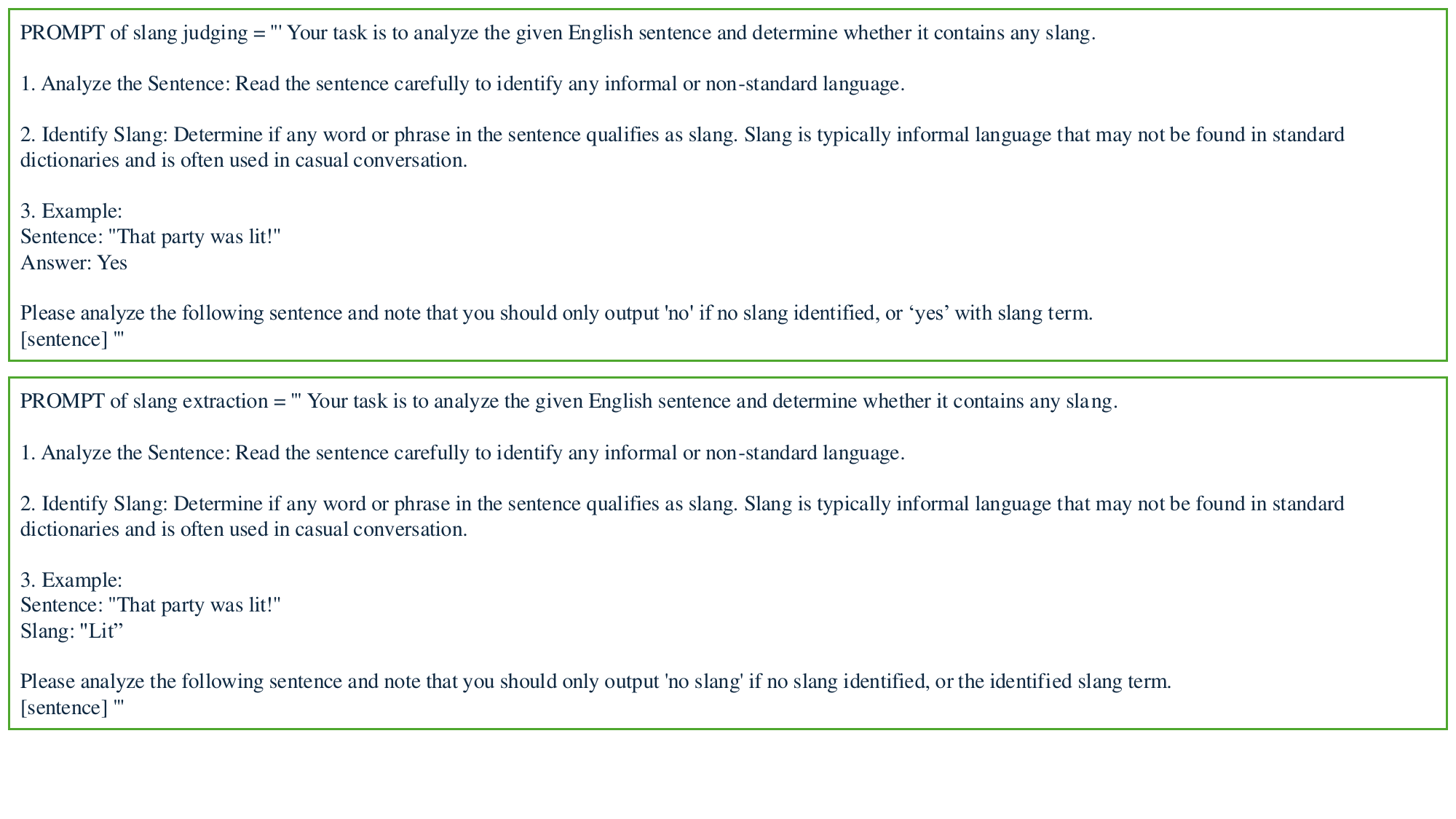}
    \caption{The prompt used in the section of Annotation Procedure.}
    \label{annotaion_app1}
\end{figure*}}

\textbf{\begin{figure*}[t]
    \centering
    \includegraphics[width=0.99\textwidth]{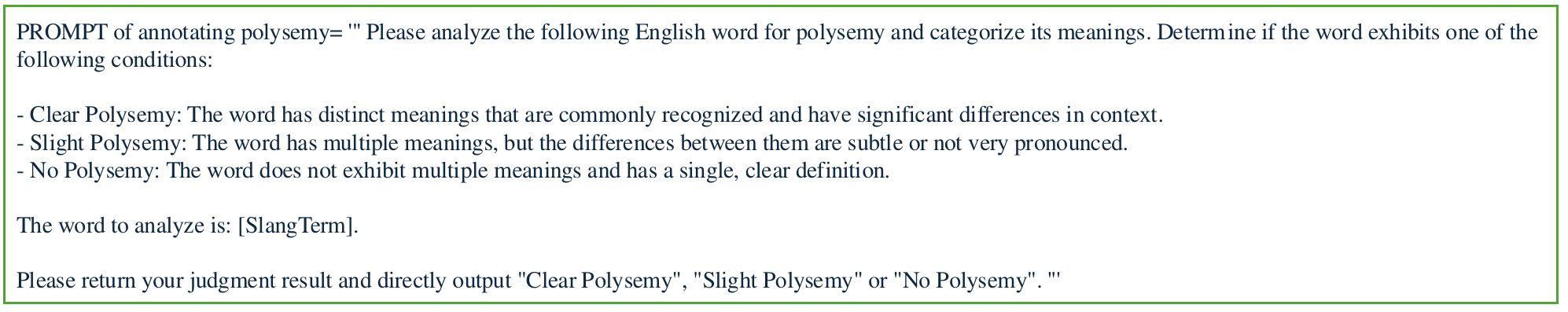}
    \caption{The prompt used in the section of Annotation Procedure.}
    \label{annotaion_app2}
\end{figure*}}

\textbf{\begin{figure*}[t]
    \centering
    \includegraphics[width=0.99\textwidth]{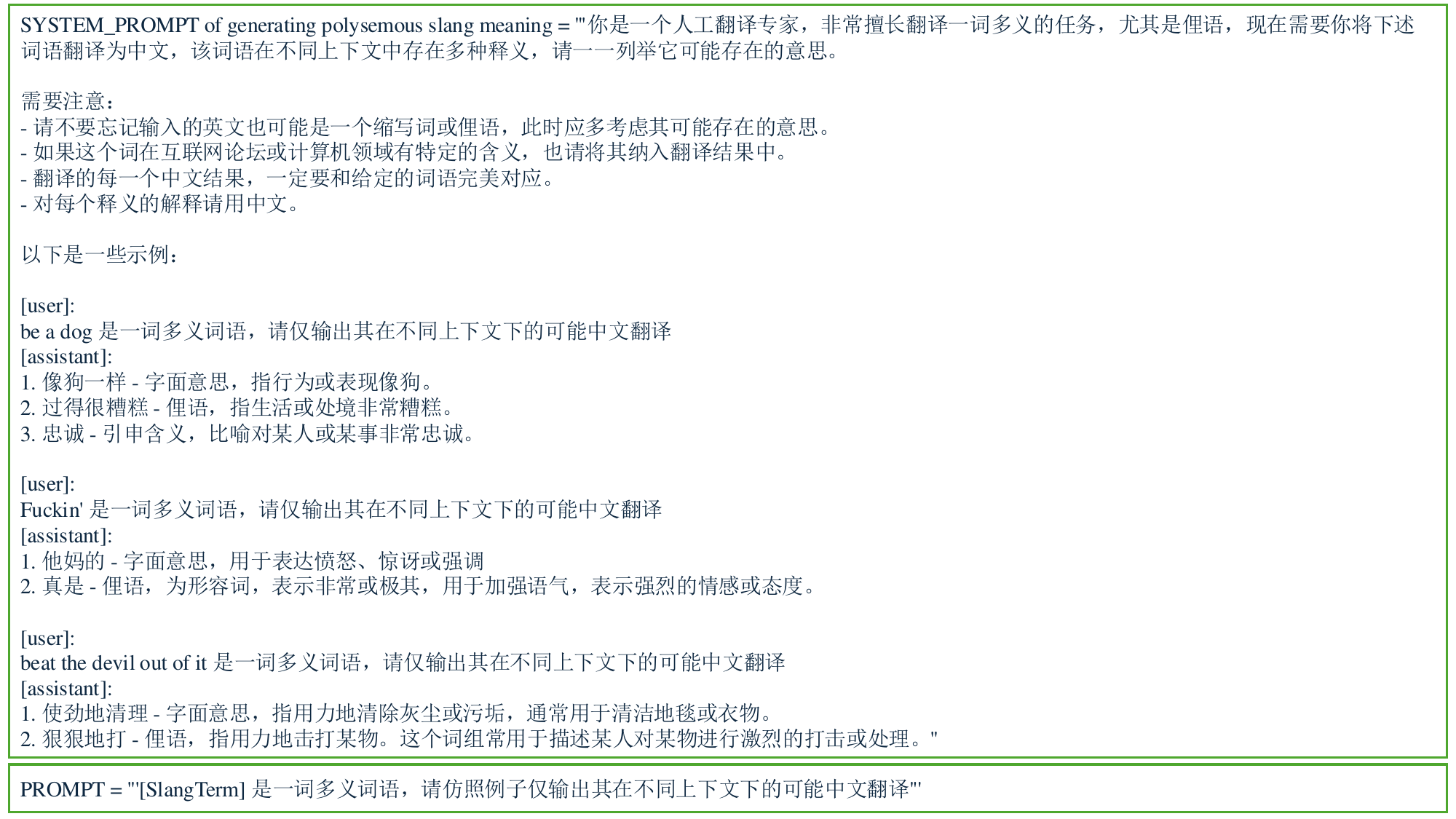}
    \caption{The prompt used in the section of Annotation Procedure.}
    \label{annotaion_app3}
\end{figure*}}

\textbf{\begin{figure*}[t]
    \centering
    \includegraphics[width=0.99\textwidth]{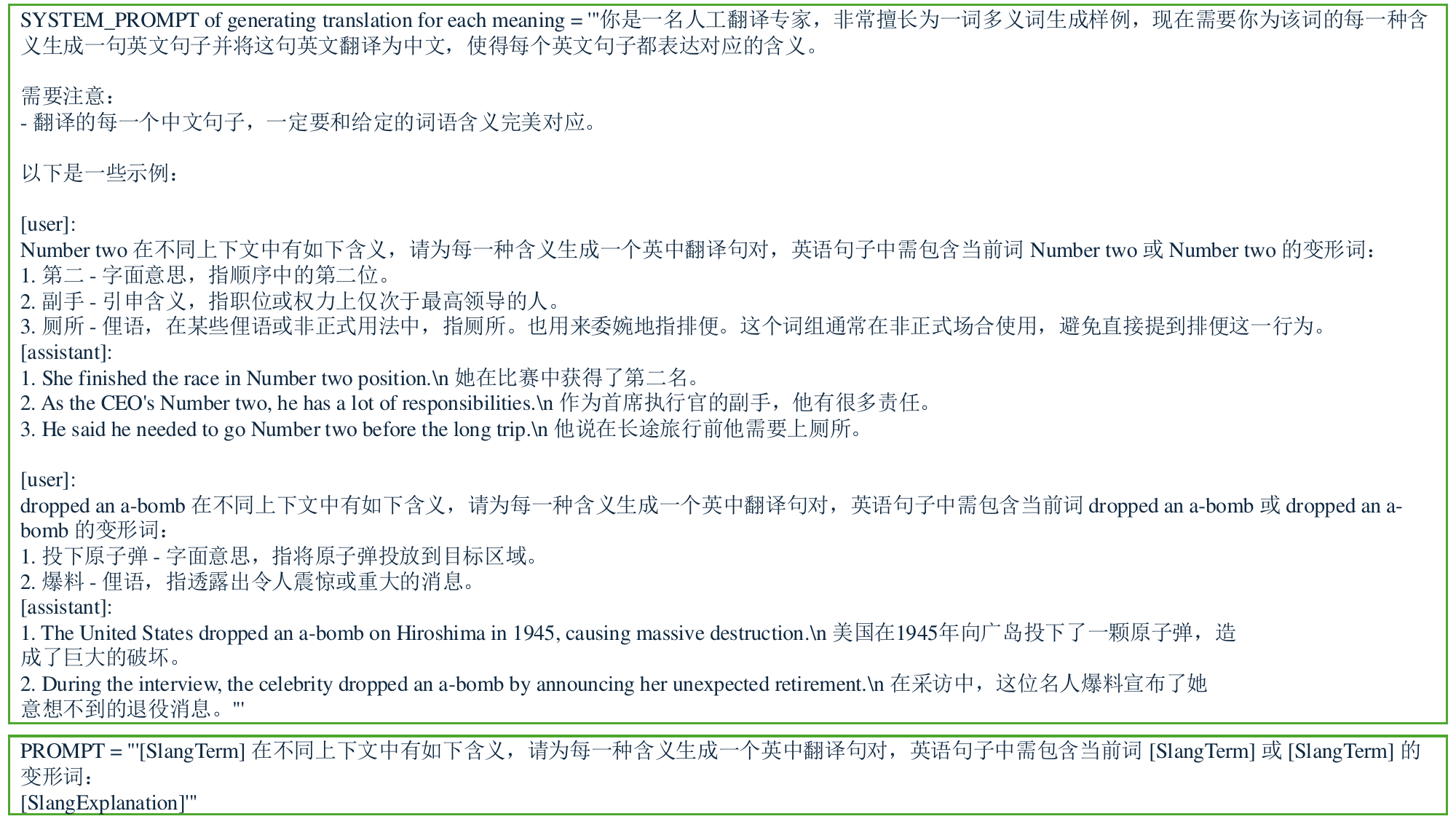}
    \caption{The prompt used in the section of Annotation Procedure.}
    \label{annotaion_app4}
\end{figure*}}

\end{document}